%% file: main-jmlr.tex
\documentclass[pmlr,twocolumn,10pt]{jmlr} 

\usepackage{booktabs}
\usepackage{siunitx}

\usepackage{dsfont}
\usepackage{algpseudocode}
\usepackage{algorithm}
\usepackage{enumitem}
\usepackage{amsmath}
\usepackage{breqn}
\usepackage{multirow}
\usepackage{adjustbox}
\usepackage{float}
\usepackage{multirow}
\usepackage{lmodern}
\usepackage{siunitx}
\usepackage{booktabs}
\usepackage{etoolbox}
\usepackage{bm}

\renewrobustcmd{\bfseries}{\fontseries{b}\selectfont}
\renewrobustcmd{\boldmath}{}
\newrobustcmd{\B}{\bfseries}

\newcommand{\descr}[1]{\vspace{0.1cm}\noindent\textit{#1}}

\newcommand{\pr}{\text{P}}

\theorembodyfont{\upshape}
\theoremheaderfont{\scshape}
\theorempostheader{:}
\theoremsep{\newline}

\title[]{Practical Challenges in Differentially-Private Federated Survival Analysis of Medical Data}

\author{%
\Name{Shadi Rahimian}\Email{shadi.rahimian@cispa.saarland}\\
\addr CISPA Helmholtz Center for Information Security, Germany 
\AND
\Name{Raouf Kerkouche}\Email{raouf.kerkouche@cispa.de}\\
\addr CISPA Helmholtz Center for Information Security, Germany 
\AND
\Name{Ina Kurth}\Email{ina.kurth@dkfz-heidelberg.de}\\
\addr DKFZ German Cancer Research Center, Germany 
\AND
\Name{Mario Fritz}\Email{fritz@cispa.de}\\
\addr CISPA Helmholtz Center for Information Security, Germany 
}

\begin{document}

\maketitle

\begin{abstract}
Survival analysis or time-to-event analysis aims to model and predict the time it takes for an event of interest to happen in a population or an individual. In the medical context this event might be the time of dying, metastasis, recurrence of cancer, etc. Recently, the use of neural networks that are specifically designed for survival analysis has become more popular and an attractive alternative to more traditional methods. In this paper, we take advantage of the inherent properties of neural networks to federate the process of training of these models. This is crucial in the medical domain since data is scarce and collaboration of multiple health centers is essential to make a conclusive decision about the properties of a treatment or a disease. To ensure the privacy of the datasets, it is common to utilize differential privacy on top of federated learning. Differential privacy acts by introducing random noise to different stages of training, thus making it harder for an adversary to extract details about the data. However, in the realistic setting of small medical datasets and only a few data centers, this noise makes it harder for the models to converge. 
To address this problem, we propose DPFed-post which adds a post-processing stage to the private federated learning scheme. This extra step helps to regulate the magnitude of the noisy average parameter update and easier convergence of the model. For our experiments, we choose 3 real-world datasets in the realistic setting when each health center has only a few hundred records, and we show that DPFed-post successfully increases the performance of the models by an average of up to $17\%$ compared to the standard differentially private federated learning scheme.



\end{abstract}


\input{introduction}
\input{background}
\input{approach}

\input{conclusion}


\section{Acknowledgments}

This work is partially funded by the Helmholtz Association within
the project “Trustworthy Federated Data Analytics” (TFDA)
(funding number ZT-I-OO1 4).

\bibliography{references}

\appendix

\section{Algorithms}\label{apd:first}

\begin{algorithm}[h]
	\caption{\texttt{\\Federated Learning (StdFed)}.\\ $N$ total clients, local mini-batch size $B$, local epochs $E$, communication rounds $T_{cl}$ and learning rate $\eta$.}
	\label{algo:fed}
	\begin{flushleft}
	 Initialize $\mathbf{w}_0$ and send the model to clients\\
    \textbf{for} $r = 1,...T_{cl}$\\
    \hspace*{\algorithmicindent} Select $K$ clients randomly \\
    \hspace*{\algorithmicindent}\textbf{for} each selected client $k=1,...,K$\\
    \hspace*{\algorithmicindent}\hspace*{\algorithmicindent}$\textbf{w}^{r}_k\leftarrow\text{ClientUpdate}(k, \textbf{w}^{r-1})$\\
    \hspace*{\algorithmicindent}$\mathbf{w}^r \leftarrow \mathbf{w}^{r-1}+\frac{\sum_{k=1}^K (\mathbf{w}_k^r - \mathbf{w}^{r-1})}{K}$\\

    \textbf{ClientUpdate}$(k, \mathbf{w})$\\
    \hspace*{\algorithmicindent}for client $k$\\
    \hspace*{\algorithmicindent}\textbf{for} $i=1,..., E$\\
    \hspace*{\algorithmicindent}\hspace*{\algorithmicindent}\textbf{for} local batches $b$\\
    \hspace*{\algorithmicindent}\hspace*{\algorithmicindent}\hspace*{\algorithmicindent}$\textbf{w}\leftarrow \textbf{w} - \eta\nabla l(b;\textbf{w})$\\
    \hspace*{\algorithmicindent}\textbf{return} $\mathbf{w}$ to server\\
	\end{flushleft}
\end{algorithm}

\begin{algorithm}[h]
	\caption{\texttt{\\Differentially Private Federated Learning with post processing (DPFed)}.\\ $N$ total clients, local mini-batch size $B$, local epochs $E$, communication rounds $T_{cl}$, learning rate $\eta$, sensitivity $S$ and post-processing parameter $P$.}
	\label{algo:DPfed}
	\begin{flushleft}
	 Initialize $\mathbf{w}_0$ and send the model to clients\\
    \textbf{for} $r = 1,...T_{cl}$\\
    \hspace*{\algorithmicindent} Select $K$ clients randomly \\
    \hspace*{\algorithmicindent}\textbf{for} each selected client $k=1,...,K$\\
    \hspace*{\algorithmicindent}\hspace*{\algorithmicindent}$\textbf{w}^{r}_k\leftarrow\text{ClientUpdate}(k, \textbf{w}^{r-1})$\\
    \hspace*{\algorithmicindent}\hspace*{\algorithmicindent}$\Delta\textbf{w}^{r}_k\leftarrow \mathbf{w}_k^r - \mathbf{w}^{r-1} $\\
    \hspace*{\algorithmicindent}\hspace*{\algorithmicindent}$\Delta\hat{\textbf{w}}^{r}_k\leftarrow \Delta{\textbf{w}}^{r}_k/\max\left(1, \frac{||\Delta \textbf{w}^{r}_k||_2}{S}\right)$\\

    \hspace*{\algorithmicindent}$\Delta{\textbf{w}}^{r} \leftarrow \frac{\sum_{k=1}^K \Delta\hat{\textbf{w}}^{r}_k + \mathcal{G}(0,S\sigma\mathbf{I})}{K}$\\
    
    \hspace*{\algorithmicindent}$\mathbf{w}^r \leftarrow \mathbf{w}^{r-1}+\Delta{\textbf{w}}^{r}$\\

    \textbf{ClientUpdate}$(k, \mathbf{w})$\\
    \hspace*{\algorithmicindent}for client $k$\\
    \hspace*{\algorithmicindent}\textbf{for} $i=1,..., E$\\
    \hspace*{\algorithmicindent}\hspace*{\algorithmicindent}\textbf{for} local batches $b$\\
    \hspace*{\algorithmicindent}\hspace*{\algorithmicindent}\hspace*{\algorithmicindent}$\textbf{w}\leftarrow \textbf{w} - \eta\nabla l(b;\textbf{w})$\\
    \hspace*{\algorithmicindent}\textbf{return} $\mathbf{w}$ to server\\
	\end{flushleft}
\end{algorithm}

\section{Results}\label{apd:second}

Results for METABRIC and SUPPORT datasets according to Section.~\ref{sec:setup} and Section.~\ref{sec:observations}. The arrows indicate if a lower or a higher value of the metric indicates better utility of the model. 
To compare our proposed post-processing method with the vanilla DP federated learning (DPFed), we have made numbers bold only when our method offers an advantage. 

\begin{table*}[htbp]\footnotesize
\floatconts
  {tab:metabric}
  {\caption{METABRIC dataset}}
    {\begin{tabular}{c|c|cc|cc|cc}
    \toprule
    &&\multicolumn{2}{c|}{}&\multicolumn{2}{c|}{}&\multicolumn{2}{c}{}\\
    & &\multicolumn{2}{c|}{Non-Private}&\multicolumn{2}{c|}{$(\epsilon=5.4, \delta=10^{-3})$}&\multicolumn{2}{c}{$(\epsilon=8.9, \delta=10^{-3})$} \\ 

    Metric&Model&Centralized& StdFed& DPFed & $\text{DPFed}_{\text{post}}$ &DPFed  &$\text{DPFed}_{\text{post}}$\\
    \hline
         & DeepHit&  $0.68\pm0.17$& $0.65\pm0.02$& $0.51\pm0.06$&\bm{$0.52\pm0.03$}&$0.53\pm0.02$&$\bm{0.57\pm0.04}$\\
      C-index  &CoxPH&$0.64\pm0.12$& $0.65\pm0.01$&$0.50\pm0.08$&\bm{$0.52\pm0.03$}&$0.60\pm0.02$&\bm{$0.63\pm0.04$}\\
         $\uparrow$&CoxCC&$0.62\pm0.01$& $0.62\pm0.02$&$0.50\pm0.03$&\bm{$0.51\pm0.07$}&$0.53\pm0.06$&\bm{$0.55\pm0.08$}\\
         &CoxTime&$0.63\pm0.12$& $0.64\pm0.01$&$0.53\pm0.05$&$0.51\pm0.05$&$0.56\pm0.01$&$0.55\pm0.06$\\
         \hline
         &DeepHit&$0.16\pm0.01$&$0.17\pm0.01$&$0.21\pm0.01$&\bm{$0.18\pm0.01$}&$0.21\pm0.01$&\bm{$0.18\pm0.01$}\\
      IBS  &CoxPH&$0.17\pm0.01$&$0.18\pm0.01$&$0.20\pm0.01$&\bm{$0.19\pm0.01$}&$0.19\pm0.01$&\bm{$0.18\pm0.01$}\\
        $\downarrow$ &CoxCC&$0.17\pm0.01$&$0.17\pm0.01$&$0.19\pm0.01$&$0.20\pm0.01$&$0.19\pm0.01$&\bm{$0.18\pm0.01$}\\
         &CoxTime&$0.17\pm0.01$&$0.17\pm0.01$&$0.21\pm0.01$&\bm{$0.20\pm0.01$}&$0.20\pm0.01$&\bm{$0.19\pm0.01$}\\
         \hline
         &DeepHit&$0.50\pm0.02$&$0.52\pm0.01$&$0.67\pm0.02$&\bm{$0.54\pm0.01$}&$0.62\pm0.02$&\bm{$0.55\pm0.02$}\\
        NIBLL &CoxPH&$0.50\pm0.02$&$0.48\pm0.01$&$0.57\pm0.03$&$0.58\pm0.03$&$0.55\pm0.01$&$0.55\pm0.03$\\
      $\downarrow$ &CoxCC&$0.51\pm0.02$&$0.52\pm0.01$&$0.57\pm0.02$&$0.57\pm0.03$&$0.56\pm0.01$&\bm{$0.55\pm0.03$}\\
         &CoxTime&$0.51\pm0.01$&$0.51\pm0.01$&$0.60\pm0.05$&\bm{$0.58\pm0.03$}&$0.61\pm0.02$&\bm{$0.53\pm0.02$}\\
         \bottomrule
    \end{tabular}}
\end{table*}

\begin{table*}\footnotesize
\floatconts
  {tab:support}
  {\caption{SUPPORT dataset}}
    {\begin{tabular}{c|c|cc|cc|cc}
    \toprule
    &&\multicolumn{2}{c|}{}&\multicolumn{2}{c|}{}&\multicolumn{2}{c}{}\\
    & &\multicolumn{2}{c|}{Non-Private}&\multicolumn{2}{c|}{$(\epsilon=5.4, \delta=10^{-3})$}&\multicolumn{2}{c}{$(\epsilon=8.9, \delta=10^{-3})$} \\ 
    Metric&Model&Centralized& StdFed& DPFed & $\text{DPFed}_{\text{post}}$ &DPFed  &$\text{DPFed}_{\text{post}}$\\
    \hline
         &DeepHit& $0.61\pm0.01$& $0.59\pm0.02$& $0.47\pm0.02$&$0.49\pm0.01$&$0.49\pm0.01$&\bm{$0.51\pm0.01$} \\
      C-index  &CoxPH&$0.61\pm0.01$& $0.61\pm0.01$& $0.50\pm0.02$&\bm{$0.51\pm0.02$}&$0.51\pm0.03$&\bm{$0.53\pm0.02$}\\
         $\uparrow$&CoxCC&$0.58\pm0.01$& $0.61\pm0.02$&$0.51\pm0.04$&$0.50\pm0.02$&$0.53\pm0.03$&$0.53\pm0.01$\\
         &CoxTime&$0.60\pm0.13$&$0.59\pm0.01$&$0.54\pm0.01$&$0.51\pm0.01$&$0.49\pm0.01$&\bm{$0.53\pm0.02$}\\
         \hline
         &DeepHit&$0.19\pm0.01$&$0.23\pm0.02$&$0.31\pm0.03$&\bm{$0.25\pm0.02$}&$0.26\pm0.02$&\bm{$0.25\pm0.01$}\\
      IBS  &CoxPH&$0.19\pm0.01$&$0.19\pm0.01$&$0.22\pm0.01$&\bm{$0.21\pm0.01$}&$0.21\pm0.01$&$0.21\pm0.01$\\
        $\downarrow$ &CoxCC&$0.20\pm0.01$&$0.19\pm0.01$&$0.22\pm0.02$&$0.22\pm0.01$&$0.21\pm0.01$&$0.21\pm0.01$\\
         &CoxTime&$0.20\pm0.01$&$0.20\pm0.01$&$0.25\pm0.03$&\bm{$0.21\pm0.01$}&$0.23\pm0.01$&\bm{$0.21\pm0.01$}\\
         \hline
         &DeepHit&$0.57\pm0.01$&$0.67\pm0.06$&$0.97\pm0.90$&\bm{$0.70\pm0.04$}&$0.75\pm0.06$&\bm{$0.73\pm0.03$}\\
        NIBLL &CoxPH&$0.56\pm0.01$&$0.56\pm0.01$&$0.64\pm0.03$&\bm{$0.61\pm0.01$}&$0.62\pm0.03$&\bm{$0.60\pm0.01$}\\
      $\downarrow$ &CoxCC&$0.58\pm0.01$&$0.57\pm0.01$&$0.63\pm0.04$&$0.63\pm0.01$&$0.60\pm0.01$&\bm{$0.59\pm0.01$}\\
         &CoxTime&$0.58\pm0.02$&$0.58\pm0.01$&$0.71\pm0.08$&\bm{$0.60\pm0.01$}&$0.66\pm0.03$&\bm{$0.61\pm0.02$}\\
         \bottomrule
    \end{tabular}}
\end{table*}

\end{document}

%% file: introduction.tex
\section{Introduction} 
\label{sec:intro}
Survival analysis or event history analysis is an old branch of statistics dating back to the 17th century~\citep{history}. The goal of survival analysis is to predict the time an event of interest occurs. This event can be the time of death of a patient\citep[e.g.][]{goldhirsch1989costs,brenner2002long}, the time of default for a bank customer\citep[e.g.][]{baesens2005neural, dirick2017time, stepanova2002survival, laitinen2005survival}, time of unsubscribing from an online service\citep[e.g.][]{mishachandar2018predicting, lu2002predicting, lee2018game}, time until a mechanical system fails\citep[e.g.][]{hanson2004artificial, styc2016predicting} and so on. The assumption is that for each population, there is a mapping from the observed features of the individuals and their time of event. Survival analysis tools are used to learn this mapping from past data that have experienced the event and generalize to the whole population. 

As for any predictive model, it is important to have enough data points to achieve generalization. This is especially of great significance in the case of medical data with many outliers and complicated patterns. However, data is sparse and usually the centers that collect these data (e.g. hospitals or banks) are reluctant or not allowed to share their data with each other due to privacy and security reasons.

This calls for federated learning in the case of survival analysis. Federated learning~\cite{mcmahan2017communication} grants the data owners the possibility of keeping their data locally, but participate in training of a global model, jointly, with other data owners. This is achievable by choosing a  trusted central server and a machine learning model that is trained locally by each participant. The role of the central server is to iteratively collect these trained models, update the global model based on the updated local models and share the updated global model with participants. Despite the obvious benefits of learning survival models in a federated setting, there has been no study on the practical effects of federation on survival analysis tools.

Although federated learning provides a scheme in which no direct sharing of data is required, there is still the risk of information leakage through parameters/output of the model. It has been shown that an adversary can carry out successful reconstruction attacks~\citep[][]{ZhuLH19, geiping2020inverting}, membership inference attacks~\citep[][]{NasrSH19,melis2019exploiting} and feature leakage attacks~\citep[][]{melis2019exploiting}. 

One solution for protecting the privacy of these models and their training data is to utilize differential privacy (DP)~\citep[][]{Dwork2014book}. DP offers privacy through a noise addition mechanism and is based on rigorous theoretical guarantees against data information leakage. In federated learning, DP can be directly applied on the clients side such that any record in any dataset of any client is protected~\citep[record-level DP e.g. in][]{truex2020ldp, kerkoucheprivacy} or it can be applied on the updates that are shared by each client with the central server such that the client's dataset as a whole is protected.~\citep[client-level DP e.g. in][]{geyer2017differentially}. The latter is preferable as it provides tighter privacy gurantees. 


In this paper, we focus on real-world challenges of DP federated learning of survival models for medical data, where data and also the number of clients that participate in the federated learning is limited. Here, the noise of client-level DP - which scales proportional to the client sampling probability - is so large that it prevents the model from converging. To tackle this problem, we take advantage of the post-processing property of differential privacy and make an additional step of clipping the noisy average update of clients by a reasonable value. This method can be viewed as using a regularization for the learning rate of the global model. 


Our contributions are:
\begin{itemize}
\item We evaluate the effect of federation of survival models on real-world datasets
\item We discuss what the challenges would be in federation of these models for the realistic setting when a few clients each possessing only a few hundred data points wish to collaborate. 
\item We successfully apply a post-processing step for client-level differentially-private federated learning and observe that the performances of the model are consistently improved over different datasets and models when this trick is used.
\end{itemize}


%% file: background.tex
\section{Background}
\label{sec:background}
In this section, we will give an overview of all the necessary concepts to understand our approach. We first describe survival analysis in a more formal way and introduce some traditional methods that are used for time-to-event prediction. We explain why in the medical setting it is important that multiple parties collaborate and share their data such that a richer, more generalizable model can be learned.

We then introduce federated learning which is a well-motivated and popular solution to collaborative training of models. By using federated learning, different parties each holding their own dataset can jointly train a unified model. 

Naturally, collaborative training of a model is prone to privacy breaches. So at last we give a brief overview of differential privacy as a privacy-preserving method that can be deployed on top of federated learning.  

\subsection{Survival Analysis}
\label{sec:bg-survival}
Survival analysis is the collection of the tools for analysing and predicting the time duration until a specific event happens. For example, in the case of clinical data, this event can be the time of death for the patients. \\
The \textit{survival function} $S(t)$ and the \textit{cumulative incidence function} (CIF) $F(t)$ can be defined as follows:
\begin{eqnarray}
    S(t) = \pr(T>t) = 1 - F(t)
\end{eqnarray}
where $\pr(T>t)$ indicates the probability that the event of interest $T$ happens after time $t$. 
So the survival function is interpreted as the probability of the event happening after time $t$ and the complementary CIF is the probability of the event happening before or at time $t$. We can also define the \textit{hazard rate} $h(t)$:
\begin{eqnarray}
    h(t) = \lim_{\Delta t\rightarrow 0}\frac{1}{\Delta t}\pr(t\leq T < t+\Delta t | T \geq t)
\end{eqnarray}
Thus the survival function can be calculated from the hazard rate by:
\begin{equation}
S(t) = \exp[-\int_0^t h(s)ds] = \exp[- H(t)],
\end{equation}
where $H(t)$ is called the cumulative hazard. We can observe that having one of these functions is adequate to calculate the others, but each offers a unique perspective into the survival status of the data.

For real-world data, the event we are interested in, e.g. death or recovery, may not be observed for all data points. This can happen when, for example, the patient does not participate in the follow-ups, or they die due to a cause unrelated to the original study, or simply when they have not yet experienced the event of interest at the last follow-up of the study. These data points are said to be \textit{right-censored}. So the time of event is chosen as $T = \min\{T^*, C\}$, where $T^*$ is the true time of event for the individual and $C$ is the time of censoring.
Survival data usually comes in the form of:
\begin{eqnarray}
\mathcal{X} =\{\mathbf{x}_i, y_i\}_{i=1}^N =\{\mathbf{x}_i, T_i, E_i\}^N_{i=1}
\end{eqnarray}
for $N$ individuals. Here, $\mathbf{x}_i$ and $y_i$ are the vector of covariates or features and the label for data point $i$, respectively. The features vector can, for example, contain information about the age, gender, medical history, tumor size, etc. for medical data. $y_i$ can further be decomposed into the tuple of $\{T_i, E_i\}$ where $T_i$ is the time of event for the event type $E_i$ for individual $i$. It is customary to set $E = 0$ for individuals that are right-censored. 

Given this set $\mathcal{X}$, one way to calculate the survival functions is by \textit{Cox Proportional Hazards} \citep{cox1972regression} model:
\small
\begin{eqnarray}
\label{eq:cox}
h(t|\mathbf{x}) = h_0(t)g(\mathbf{x}) \qquad \text{where} \qquad g(\mathbf{x}) = {\beta}^T \mathbf{x} 
\end{eqnarray}
The main assumption of the Cox proportional hazards model is that the features $\mathbf{x}$ are independent and a linear combination of them and a non-parametric \textit{baseline hazard} $h_0(t)$ are sufficient to model the data. To fit this model, first the coefficients ${\beta}$ are found by maximizing the partial log likelihood of the model:
\small
\begin{eqnarray}
\label{eq:likelihood-cox}
L = \Pi _i \left(\frac{\exp [g(\mathbf{x}_i)]}{\sum_{j \in R_i} \exp [g(\mathbf{x}_j)]}   \right)^{E_i}
\end{eqnarray}
where $R_i$ is the risk set of all individuals at risk at time $T_i$. 

In the second step, the baseline hazard is estimated for the parameters found in the first step.

The \textit{proportionality} in the name of this model comes from the fact that for two data points $\mathbf{x}_i$ and $\mathbf{x}_j$ the hazard ratio remains constant over time:
\small
\begin{eqnarray}
\frac{h(t|\mathbf{x}_i)}{h(t|\mathbf{x}_j)} = \frac{h_0(t)g(\mathbf{x}_i)}{h_0(t)g(\mathbf{x}_j)} = \frac{g(\mathbf{x}_i)}{g(\mathbf{x}_j)}
\end{eqnarray}

 To summarize, the main goal of survival analysis is to model the relationship between the vector of features of the data points and the time that an event of interest happens. This model is fitted on a set of data points that we assume is representative of the statistical properties of a population. The more datapoints are used to fit the model, the more generalizable the model will be, as each datapoint adds more information about the properties of the population. 

Unfortunately, the collection of data is expensive and time consuming for the health centers, and the number of patients they receive is also limited. This means that by using only the data of one health center, we would have a high bias and the fitted model would not be well-generalizable. As a solution for this problem, we look at the federated learning for survival models, in the next section. Federated learning allows multiple data-holders to jointly learn a model over the union of their data. 

\subsection{Federated Learning}
\label{sec:bg-fedl}
One of the major issues for modeling survival data is the lack of enough training data. Oftentimes many separate data owners have access to limited amounts of data and due to reasons, such as privacy and security, are unwilling to simply share their data with each other to learn a richer model. This is where federated learning \cite{mcmahan2017communication, shokri2015privacy} becomes important. Federated learning is the concept of distributing the learning process of machine learning models among several data owners, without the need to access their local data. 

In this paper we use deep neural networks designed for survival analysis as the model that we wish to federate. In general to train a deep neural network, an objective function is optimized over the parameters of the model, $\mathbf{w}$, for a training dataset which contains $n$ data points:
\small
\begin{eqnarray}
\min_{\mathbf{w}}f(\mathbf{w})\ \text{where}\  
f(\mathbf{w}) = \frac{1}{n}\sum_{i=1}^n\mathcal{L}(\mathbf{x}_i, y_i;\mathbf{w})
\end{eqnarray}
where $\mathcal{L}(\mathbf{x}_i, y_i;\mathbf{w})$ is the loss function for $i$-th data point with label $y_i$, and the training is done iteratively over the dataset. Therefore by defining the appropriate  loss function for survival analysis tasks, we can train the network to capture the behavior of the stochastic processes from the training set.

In federated learning a central server holds a global model which is also shared with all the clients. Clients train this model locally and only send their updated model parameters to the central server. The server updates the parameters of the global model with a weighted average of the local parameters and re-shares the updated global model with the clients. To reduce the cost of communication and computation, usually at each round, only a fraction of clients, $C = K/N$,  are chosen randomly to train their local model. Here $C$ is the sampling probability of clients, $K$ is the number of clients chosen out of a total of $N$ clients at each round:
\small
\begin{eqnarray}
\label{eq:fedgeneric}
\mathbf{w}^r \leftarrow \mathbf{w}^{r-1}+\frac{\sum_{k=1}^K (\mathbf{w}_k^r - \mathbf{w}^{r-1})}{K}
\end{eqnarray}
where $\mathbf{w}_k^r$ is the model parameters for client $k$ after the $r$-th round of communication with the central server. These steps are done for a total number of $T_{cl}$ communication rounds. The local parameters $\mathbf{w}_k^r$ are computed by each client using the standard training algorithms, such as stochastic gradient decent~\citep{robbins1951stochastic, kiefer1952stochastic}. Here, it is assumed that each client possesses $n_k$ training data and total number of data points can be calculated as $n = \sum_{k=1}^K n_k$. 

Although Federated Learning improves privacy by design, model parameters can leak information about the training data. Indeed, \cite{ZhuLH19,idlg,geiping2020inverting} presented some attacks that allow an adversary to reconstruct some training data samples of some entities. \cite{NasrSH19,melis2019exploiting} define a membership attack that allows to infer if a particular record is included in the data of a specific entity. Similarly, ~\cite{melis2019exploiting} define an attack which aims at inferring if a group of people with a specific property, like for example skin color or ethnicity, is included in the dataset of a particular participating entity. A solution to prevent these attacks and provide theoretical guarantees in to use a privacy model called Differential Privacy \citep{Dwork2014book}. 

\subsection{Differential Privacy}
\label{sec:bg-dp}
Differential privacy (DP) allows an entity to privately release information about a dataset:  a function of a record dataset is perturbed, so that any information which can differentiate this record from the rest of the dataset is bounded~\cite{Dwork2014book}.

 \begin{definition}[Privacy loss]
 Let $\mathcal{A}$ be a privacy mechanism which assigns a value $\mathit{Range}(\mathcal{A})$ to a dataset $D$. The privacy loss of $\mathcal{A}$ with datasets $D$ and $D'$ at output $O \in \mathit{Range}(\mathcal{A})$ is a random variable $\mathcal{P}(\mathcal{A},D,D',O) = \log\frac{\Pr[\mathcal{A}(D) = O]}{\Pr[\mathcal{A}(D') = O]}$ 
 where the probability is taken on the randomness of $\mathcal{A}$.
 \label{def:ploss}
 \end{definition}

\begin{definition}[$(\epsilon,\delta)$-Differential Privacy]
A privacy mechanism $\mathcal{A}$ guarantees $(\varepsilon, \delta)$-differential privacy if for any database $D$ and $D'$, differing on at most one record, $\Pr_{O \sim \mathcal{A}(D)}[\mathcal{P}(\mathcal{A},D,D',O) > \varepsilon] \leq \delta$~\citet{Dwork2014book}.

\label{def:DP}
\end{definition}

This guarantees that an adversary, who has access to the output of $\mathcal{A}$, can draw almost the same conclusions up to $\varepsilon$ (with probability larger than $1 - \delta$) about any record no matter if it is included in the input of $\mathcal{A}$ or not.

\descr{Moments Accountant.}  Differential privacy maintains composition; the privacy guarantee of the  $k$-fold adaptive composition  of $\mathcal{A}_{1:k} = \mathcal{A}_1, \ldots, \mathcal{A}_k$ can be computed using the moments accountant method \cite{Abadi}. In particular, it follows from Markov's inequality that $\Pr[\mathcal{P}(\mathcal{A},D,D',O) \geq \varepsilon] \leq \mathbb{E}[\exp(\lambda \mathcal{P}(\mathcal{A},D,D',O))]/\exp(\lambda\varepsilon)$ for any output $O \in \mathit{Range}(\mathcal{A})$ and $\lambda > 0$.  
$\mathcal{A}$ is $(\varepsilon, \delta)$-DP 
with $\delta = \min_{\lambda} \exp(\alpha_{\mathcal{A}}(\lambda) - \lambda \varepsilon)$, where $\alpha_{\mathcal{A}}(\lambda) = $\\ $ \max_{D,D'} \log\mathbb{E}_{O\sim \mathcal{A}(D)}[\exp(\lambda \mathcal{P}(\mathcal{A},D,D',O))]$ is the log of the moment generating function of the privacy loss. The privacy guarantee of the composite mechanism $\mathcal{A}_{1:k}$ can be calculated using that $\alpha_{\mathcal{A}_{1:k}}(\lambda) \leq \sum_{i=1}^k \alpha_{\mathcal{A}_{i}}(\lambda)$ \cite{Abadi}. \smallskip 



\descr{Gaussian Mechanism.} 
A fundamental concept of all DP sanitization techniques is the \emph{global sensitivity} of a function~\citep{Dwork2014book}.
\begin{definition}[Global $L_p$-sensitivity] 
For any function $f:\mathcal{D} \rightarrow \mathbb{R}^ n$, the $L_p$-sensitivity of $f$ is
$\Delta_p f = \max_{D, D'} || f(D)-f(D') ||_p$, 
for all $D, D'$ differing in at most one record, where $||\cdot||_p$ denotes the $L_p$-norm.\vspace*{-0.15cm}
\label{def:global_sens}
\end{definition}
\smallskip
The Gaussian Mechanism~\citep{Dwork2014book} 
consists of adding Gaussian noise to the true output of a function.
In particular, for any function $f:\mathcal{D} \rightarrow \mathbb{R}^n$, the Gaussian mechanism is defined as adding i.i.d Gaussian noise with variance $(\Delta_2 f \cdot \sigma)^2$  and zero mean to each coordinate value of  $f(D)$. Recall that the pdf of the Gaussian distribution with mean $\mu$ and variance $\xi^2$ is
$
\mathsf{pdf}_{\mathcal{G}(\mu, \xi)}(x) = \frac{1}{\sqrt{2\pi}\xi} \exp\left(-\frac{(x-\mu)^2}{2 \xi^2}\right) 
$.

In fact, the Gaussian mechanism draws vector values from a multivariate spherical (or isotropic) Gaussian distribution which is described by random variable $\mathcal{G}(f(D), \Delta_2 f \cdot \sigma\mathbf{I}_n)$, where $n$ is omitted if its unambiguous in the given context.


%% file: approach.tex
\section{Approach}
\label{sec:approach}
The most successful research in privacy-preserving federeted learning has been studied on huge datasets, containing up to millions of data records and hundreds of clients~\citep[see e.g.][]{kairouz2019advances, augenstein2019generative, mcmahan2017communication, truex2019hybrid}. However, in the realistic setting of medical application, the data is very scarce and the number of clients that are willing to collaborate, i.e. health centers, is also limited. This poses problems such as instability of results as well as sensitivity of the model to the DP noise when a privacy-preserving solution is needed. 

In this section, we first define the privacy model and the layer we choose to add differential privacy to, then discuss the problems that this noise would cause in the case of small datasets. We propose our post-processing technique which does not change the DP guarantees, however, acts as a regularization on the learning rate thus stabilizing the convergence of the model during the federated training. 

Finally, we present our experiments using 4 different survival deep neural networks and compare the results we obtain by federation, differentially private federation as well as applying the post-processing step.  

\begin{algorithm}[h]
\small
	\caption{\texttt{\\Differentially Private Federated Learning with post processing (DPFed-post)}.\\ $N$ total clients, local mini-batch size $B$, local epochs $E$, communication rounds $T_{cl}$, learning rate $\eta$, sensitivity $S$ and post-processing parameter $P$.}
	\label{algo:DPfedpost}
	\begin{flushleft}
	 Initialize $\mathbf{w}_0$ and send the model to clients\\
    \textbf{for} $r = 1,...T_{cl}$\\
    \hspace*{\algorithmicindent} Select $K$ clients randomly \\
    \hspace*{\algorithmicindent}\textbf{for} each selected client $k=1,...,K$\\
    \hspace*{\algorithmicindent}\hspace*{\algorithmicindent}$\textbf{w}^{r}_k\leftarrow\text{ClientUpdate}(k, \textbf{w}^{r-1})$\\
    \hspace*{\algorithmicindent}\hspace*{\algorithmicindent}$\Delta\textbf{w}^{r}_k\leftarrow \mathbf{w}_k^r - \mathbf{w}^{r-1} $\\
    \hspace*{\algorithmicindent}\hspace*{\algorithmicindent}$\Delta\hat{\textbf{w}}^{r}_k\leftarrow \Delta{\textbf{w}}^{r}_k/\max\left(1, \frac{||\Delta \textbf{w}^{r}_k||_2}{S}\right)$\\

    \hspace*{\algorithmicindent}$\Delta{\textbf{w}}^{r} \leftarrow \frac{\sum_{k=1}^K \Delta\hat{\textbf{w}}^{r}_k + \mathcal{G}(0,S\sigma\mathbf{I})}{K}$\\
    
    \hspace*{\algorithmicindent}$\Delta{\hat{\textbf{w}}}^{r} \leftarrow \Delta{\textbf{w}}^{r} / \max\left(1, \frac{||\Delta \textbf{w}^{r}||_2}{P}\right)$ \\
    
    \hspace*{\algorithmicindent}$\mathbf{w}^r \leftarrow \mathbf{w}^{r-1}+\Delta{\hat{\textbf{w}}}^{r}$\\

    \textbf{ClientUpdate}$(k, \mathbf{w})$\\
    \hspace*{\algorithmicindent}for client $k$\\
    \hspace*{\algorithmicindent}\textbf{for} $i=1,..., E$\\
    \hspace*{\algorithmicindent}\hspace*{\algorithmicindent}\textbf{for} local batches $b$\\
    \hspace*{\algorithmicindent}\hspace*{\algorithmicindent}\hspace*{\algorithmicindent}$\textbf{w}\leftarrow \textbf{w} - \eta\nabla l(b;\textbf{w})$\\
    \hspace*{\algorithmicindent}\textbf{return} $\mathbf{w}$ to server\\
	\end{flushleft}
\end{algorithm}

\subsection{Privacy Model}
\label{sec:privacy_model}
We consider an adversary, or a set of colluding adversaries, who can access any update vector sent by the server at each round of the protocol.  A plausible adversary is a participating entity, i.e. a malicious client, that wants to infer the training data used by other participants. We assume that the server is trusted.
The adversary is \emph{passive} (i.e., honest-but-curious), that is, it follows the learning protocol faithfully. 

Different privacy requirements can be considered depending on what information the adversary aims to infer. In general, private information can be inferred about:
\begin{itemize}
    \item any record (user or patient) in any dataset of any client (\emph{record-level privacy}),
    \item any client/party (\emph{client-level privacy}).
\end{itemize}

To illustrate the above requirements, suppose that several hospitals build a common model to predict the risk of death for their patients. A hospital certainly does not want other hospitals to learn the  status of any of their patients (record privacy) and perhaps not even the average status of all their patients  (client privacy).

Record-level privacy is a standard requirement used in the privacy literature, but it is weaker than client-level privacy. Indeed, client-level privacy requires to hide any information which is unique to a client including perhaps all its training data.   


We aim at developing a solution that provides \emph{client-level privacy}. For example, in the scenario of collaborating hospitals, we aim at protecting any information that is unique to each single hospital's training data.
The adversary should not be able to infer from the received model or its updates whether any client's data is involved in the federated run (up to $\varepsilon$ and $\delta$). 
We believe that this adversarial model is reasonable in many practical applications when the confidential information spans over multiple samples in the training data of a single client (e.g., the presence of a group a samples, such as people from a certain race). Differential Privacy guarantees plausible deniability not only to any groups of samples of a client but also to any client in the federated run. Therefore, any negative privacy impact on a party (or its training samples) cannot be attributed to their involvement in the protocol run.

\subsection{DPFed-post: Differentially Private 
Federated Learning with Post-Processing}


\subsubsection{Challenge}

Although DPFed generates a model which protects the dataset of each participant with a theoretical privacy guarantee (bounded by $\epsilon$), the utility of the model is generally very bad, due to the large amount of noise required by DP for a reasonable $\epsilon$ budget. Indeed, this noise prevents generally the convergence of the model or highly decreases its performances.

Recent works show that it is possible to improve the utility of a differentially private model by: reducing the sensitivity of the model's update ($S$) and/or increasing the value to noise level \citep{kerkouchecompression,kerkoucheconstrained} or even by local adaptation \cite{yu2020salvaging}. Note that, the sensitivity $S$ is not the only DP parameter which governs the noise. Indeed, reducing the sampling probability $C=K/N$ is also one manner to reduce the noise for a given $\epsilon$ budget. However, in certain scenarios as ours when we have only few participants (small $N$) with small datasets, the convergence of the model before using DP can be possible only with a large sampling probability ($C$ is set to 0.5 in our case). Therefore, noise might be very large in such cases, e.g. the noise multiplier $\sigma$ is set to $\sigma={2,3}$ which result in $\epsilon={5.4, 8.9}$, respectively.

\subsubsection{Our solution}

To tackle the well-known "utility-privacy tradeoff" under challenging settings, we propose a straightforward approach called \textbf{DPFed-post}. Our scheme outperforms the standard differentially private federated learning scheme (\textbf{DPFed}) by adding one normalization step.

Indeed, in both \textbf{DPFed-post} and \textbf{DPFed}, at each round $r$ the server sends the global model $\textbf{w}^{r-1}$ to $K$ clients selected randomly, each selected client trains and sends back its model to the server. The server clips the update of each client $\Delta\textbf{w}^{r}_k$ to have a $L_2$-norm at most $S$. After that, the server sums up the clipped updates $\Delta\hat{\textbf{w}}^{r}_k$, adds the Gaussian Noise and average it to obtain the noisy update $\Delta{\hat{\textbf{w}}}^{r}$. 

The difference between \textbf{DPFed-post} and \textbf{DPFed} starts from this step. Instead of directly updating the global model with the averaged noisy update as it is the case in \textbf{DPFed}, in our scheme \textbf{DPFed-post}, $\Delta{\textbf{w}}^{r}$ is normalized to obtain $\Delta{\hat{\textbf{w}}}^{r}$ with $ L_2$-norm at most $P$. Finally, the normalized noisy update $\Delta{\hat{\textbf{w}}}^{r}$ is used to update the new global model $\textbf{w}^{r}$ (See Alg.~\ref{algo:DPfedpost} for more details).

This normalization acts exactly as decreasing the learning rate. Indeed, the large additive noise generally prevents the convergence to a good local minimum. Therefore, this normalization will slow down the learning process in order to reach better performances compared to the standard scheme \textbf{DPFed}. 

\subsubsection{Privacy analysis:}


The adversary can only access the noisy model which is sufficiently perturbed to ensure client-level differential privacy; any client-specific information that could be inferred from the model is tracked and quantified by the moments accountant, described in Section~\ref{sec:bg-dp}, as follows. 

Let $\eta_0(x|\xi) =  \mathsf{pdf}_{\mathcal{G}(0, \xi)}(x)$ and $\eta_1(x|\xi) =  (1-C) \mathsf{pdf}_{\mathcal{G}(0, \xi)}(x) + C \mathsf{pdf}_{\mathcal{G}(1, \xi)}(x)$ where $C$ is the sampling probability of a single client in a single round. Let
$
\alpha(\lambda| C) = \log\max(E_1(\lambda, \xi, C), E_2(\lambda, \xi, C)) 
$
where
$
E_1(\lambda,  \xi, C) =  \int_{\mathbb{R}}\eta_0(x|\xi, C) \cdot \left(\frac{\eta_0(x|\xi, C)}{\eta_1(x|\xi, C)}\right)^{\lambda} dx
$ and
$ E_2(\lambda,  \xi, C) = \int_{\mathbb{R}}\eta_1(x|\xi, C) \cdot \left(\frac{\eta_1(x|\xi, C)}{\eta_0(x|\xi, C)}\right)^{\lambda} dx
$.

\begin{theorem}[Privacy of DPFed-post]
\label{thm:dg_privacy}
DPFed-post is $(\min_\lambda  (T_{\mathsf{cl}}\cdot \alpha (\lambda | C)  - \log \delta) /\lambda, \delta)$-DP. 
\end{theorem}
Given a fixed value of $\delta$, $\varepsilon$ is computed numerically  as in \cite{Abadi,MironovTZ19}.

\section{Experimental Results}

\subsection{Neural Networks for Survival Analysis}
\label{sec:nn}
In this paper, we utilize 4 deep neural networks which are designed for survival analysis. As explained in Section.~\ref{sec:bg-fedl}, by defining appropriate loss functions for neural networks, they could be used to serve specific purposes:

\textbf{CoxPH:} adopted from~\citet{kvamme2019time}, also known as DeepSurv~\citep{katzman2018deepsurv} replaces $g(\mathbf{x})$ of Equation.~\ref{eq:cox} with a fully connected deep neural network. The loss for this model is defined as: 
    \small{\begin{eqnarray}
    \label{eq:losscoxph}
    \text{Loss} = \frac{1}{n} \sum_{i:E_i=1} \log \left(\sum_{j\in {R}_i} \exp [g(\mathbf{x}_j) - g(\mathbf{x}_i)] \right)
    \end{eqnarray}}
    
\textbf{CoxCC:} adopted from~\citet{kvamme2019time}, Similar to CoxPH except for the fact that the risk set $R_i$ of the loss function is approximated with a large enough set to make calculations easier.

\textbf{CoxTime:} adopted from~\citet{kvamme2019time}, similar model to CoxCC, but the proportionality assumption of the Cox hazard model is abondoed:
    \small{\begin{eqnarray}
    h(t|\mathbf{x}) &=& h_0(t)\exp[g(t, \mathbf{x})]
    \end{eqnarray}}
    so that $g(t, \mathbf{x})$ uses the time $t$ as another covariate.
    
\textbf{DeepHit:} adopted from~\citet{kvamme2019time, lee2018deephit}, this model is fundamentally different from the previous 3 models and makes no assumption about the underlying relation between the covariates. It has a 2-part loss function; the first part measures the negative log-likelihood and the second part is a ranking loss which investigates each possible pair of data and tries to sort them with respect to their time of event. The output of all the previous models were continuous over time. However, the output of DeepHit is discretized and we need to define time-bins over the experiment span of our datasets. 
\subsection{Metrics}
\label{metric}
In this section we explain 3 different performance metrics, each offering a unique perspective into measuring how well the models predict on the data (for detailed explanation cf.~\cite{kvamme2019time}). 

\subsubsection{Integrated Brier Score}
Brier score is a performance measure for binary labels and takes into account both the discrimination as well as the calibration of the model's predictions. To make our labels - which are times of events - binary, we can pick a fixed time $t$ and binarize based on whether or not the event happens before or after this time $t$. So the generalized Brier score~\cite{graf1999assessment} can be calculated as:
\begin{equation}
\small
\begin{split}
\text{BS}(t) =\frac{1}{N} \sum_{i=1}^{N}&\Bigl[\frac{\hat{S}(t|\mathbf{x}_i)^2\mathds{1}(T_i\leq t, E_i=1)}{\hat{G}(T_i)} \\+ &\frac{(1-\hat{S}(t|\mathbf{x}_i))^2\mathds{1}(T_i>t)}{\hat{G}(t)} \Bigr] 
\end{split}
\end{equation}
where $N$ is the number of data points and $\hat{G}(t)$ is the Kaplan-Meier estimate for censoring survival function $\pr(T>t)$ which helps to account also for the censored data in this metric. Note that $E_i=1$ and $E_i=0$ represent datapoints experiencing the event of interest and censoring, respectively. 

To further extend the BS$(t)$ to all the possible time values $t$, we can integrate over time:
\begin{equation}
\small
\label{eq:ibs}
\text{Integrated BS} = \text{IBS} =  \frac{1}{t_2-t_1}\int_{t_1}^{t_2}\text{BS}(s)ds
\end{equation}
which can be approximated numerically by a time grid over the test set. Note that a \textbf{lower value} of the Brier score or integrated Brier score indicates better performance of the model.

\subsubsection{Integrated Binomial Log-Likelihood}
This is also another binary classification metric that measures discrimination and calibration. In the same fashion as Brier score, we can binarized the labels and define the mean binomial log likelihood as:

\begin{equation}
\begin{split}
\text{BLL}(t) =& \frac{1}{N} \sum_{i=1}^{N} \Bigl[\frac{\log(1-\hat{S}(t|\mathbf{x}_i))\mathds{1}(T_i\leq t, E_i=1)}{\hat{G}(T_i)}  \\ 
&+\frac{\log(\hat{S}(t|\mathbf{x}_i))\mathds{1}(T_i>t)}{\hat{G}(t)}\Bigr]
\end{split}
\end{equation}

We can again integrate over different times in the same way as~\ref{eq:ibs}:
\begin{eqnarray}
\label{eq:ibll}
\text{Integrated BLL} = \text{IBLL} = \frac{1}{t_2-t_1}\int_{t_1}^{t_2}\text{BLL}(s)ds
\end{eqnarray}
A higher value of the integrated binomial log-likelihood indicates better performance of the model. To avoid reporting negative values, in our experiments we calculate the negative integrated binomial log-likelihood, so here, \textbf{lower values are preferred}.
\subsubsection{Concordance Index}
 Concordance index (C-index)~\cite{harrell1982evaluating} is one of the most commonly-used metrics in the field of survival analysis. The concordance is only concerned with the ordering of the pairs of data points in the data set, regardless of the true time of the events.\\
The time-dependent C-index, for a pair of data points $i$ and $j$ can be estimated by:
\begin{equation}
\small
\begin{split}
\label{eq:cindex}
&\text{C-index} = \pr\{\hat{F}(T_i|\mathbf{x}_i) > \hat{F}(T_i|\mathbf{x}_j)| T_i < T_j, E_i = 1\}\\
&\approx\frac{\sum_{i \neq j}A_{i,j}\mathds{1}(\hat{F}(T_i|\mathbf{x}_i) > \hat{F}(T_i|\mathbf{x}_j))}{\sum_{i\neq j}A_{i,j}}
\end{split}
\end{equation}
where $\hat{F}(t|x)$ is the estimated CIF and $A_{i,j}$ is a sorting function defined as $A_{i,j} = \mathds{1}(T_i < T_j , E_i = 1)$.
The idea is that when comparing two data points, if one experiences the event sooner than the other, at that event time, it should have a lower survival probability than the other data point. A C-index of $0.5$ indicates chance-level prediction and \textbf{higher values indicate better performance}.

\subsection{Datasets}
\label{sec:dataset}
For our experiments, we choose three real-world medical datasets which are collected over multiple years:

\textbf{Rotterdam and German Breast Cancer Study Group (GBSG):} Contains data of 2232 breast cancer patients from the Rotterdam tumor bank~\cite{foekens2000urokinase} and the German Breast Cancer Study Group (GBSG)~\cite{schumacher1994randomized}. The data is pre-processed similar to~\cite{katzman2018deepsurv} and contains 7 features and the maximum survival duration is 87 months.  

\textbf{The Molecular Taxonomy of Breast Cancer International Consortium (METABRIC):} This dataset contains gene and protein expressions of 1904 individuals~\cite{curtis2012genomic}. We use a dataset prepared  similar to~\cite{katzman2018deepsurv}. This pre-processed dataset contains 4 gene indicators (MKI67, EGFR, PGR, and ERBB2) and 5 clinical features (hormone treatment indicator, radiotherapy indicator, chemotherapy indicator, ER-positive indicator,
age at diagnosis) for each patient and is followed for a maximum of 355 months.

\textbf{Study to Understand Prognoses Preferences Outcomes
and Risks of Treatment (SUPPORT):} This dataset consists of 8873 seriously-ill adults~\cite{knaus1995support}. The dataset has 14 features (age, sex, race, number of comorbidities, presence of diabetes, presence of dementia, presence
of cancer, mean arterial blood pressure, heart rate, respiration rate, temperature, white blood cell count, serum’s
sodium, and serum’s creatinine) with a maximum survival time of 67 months. We use a pre-processed version according to~\cite{katzman2018deepsurv}.

\textbf{Data in practice:} For all our experiments, we assume that 10 hospitals/data centers would like to collaborate to train a model. This is a reasonable estimate of the number of health centers that would like to study a specific disease, as seen e.g. in~\cite{tomczak2015cancer, murphy2000description, linge2016hpv}. Furthermore, in practice, each hospital would have access to only a few hundred data~\citep[e.g.][]{mohan1996fibrocalculous, gyorffy2010online, chi2007application}. In Table~\ref{tab:data-distro}, we show for each dataset how many patients were censorded, and given that the data is split between 10 hospital, how many patients each hospital would have.  
\begin{table*}\footnotesize
\floatconts
  {tab:data-distro}
  {\caption{Details of each Datasets}}
{\scalebox{0.9}{\begin{tabular}{c|c|c|c|c|c}
\toprule
    Dataset&$\#$ features&$\#$uncensored&$\#$censored& $\#$hospitals&$\#$ training data/hospital\\
    \hline
    GBSG&7&$1272(57\%)$&$960(43\%)$&10&178\\
    METABRIC&9&$1103(58\%)$&$801(42\%)$&10&152\\
    SUPPORT&14&$6034(68\%)$&$2839(32\%)$&10&709\\
    \bottomrule
\end{tabular}}}
\end{table*}

\subsection{Setup}
\label{sec:setup}
In this section, we give an overview of the technical details of our experiments in both centralized and federated setting. As mentioned in Section~\ref{sec:nn}, the output of all the models except DeepHit are continuous. For this reason we define time-bins with duration of $\sim 12$ months for all of our datasets when deploying DeepHit. We use pycox package \footnote{\url{https://github.com/havakv/pycox}} to construct all our models.

\textbf{Centralized}: To have a baseline to compare the effect of federation of the models, we first need to study their performance in a centralized setting. Datasets are randomly split into $80\%$ training and $20\%$ test data. To account for the effect of this random splitting, we run each experiment 5 times and report the average value and standard deviation. 
For all our experiments, the structure of all models are in the form of $[\text{input}, 32, 32, \text{output}]$ fully-connected nodes. We use \textsc{AdamOptimizer} with a learning rate of $10^{-4}$ to train these models and use an early stopping criterion to achieve the best performance. 

\textbf{Federated}: For all of the federated experiments, we again randomly split the data into $80\%$ training and $20\%$ test set. The training set is then randomly split into $N=10$ equal-sized parts to be assigned to each client. Here, we also run the experiments 5 times with different random splittings and report the average performance and the standard deviation.

The same neural network structure and optimizer as the centralized setting is used by clients. At each round of communication, the selected clients train their model for 50 local epochs. The training is done for a total of $T_{cl}=50$ communication rounds. 

We start our experiments in a centralized settings. In the second stage, we study the effect of standard federation (StdFed, Algorithm.~\ref{algo:fed} in Appendix A) with no privacy protocol, on the performance of the model. We then apply client-level differential privacy (DPFed, Algorithm.~\ref{algo:DPfed} in Appendix A) to this federated scheme. Finally, we utilize our post-processing solution (DPFed-post, Algorithm.~\ref{algo:DPfedpost}) on top of the client-level DP federated learning pipeline.

\textbf{Centralized to StdFed}: The first step of our experiments is to study the effect of federation of deep survival models over 10 hospitals. To pick a suitable client sampling probability $C$, we ran our experiments for $C=\{0.1, 0.2, 0.3, 0.4, 0.5, 0.6, 0.7, 0.8, 0.9, 1.0\}$ and chose the lowest value that results in a comparable performance of the federated model after 50 rounds and the centralized setting. We pick the lowest acceptable $C$ since later when DP is applied, the lower value of client sampling probability would lead to better privacy guarantees.

\textbf{StdFed to DPFed}: Many hyper-parameters play a role when applying DP on federated learning algorithm. The parameter update clipping threshold $S$, the Gaussian noise multiplier $\sigma$, the client sampling probability $C$, total number of communication rounds $T_{cl}$, and $\delta$ effect the performance of the model as well as the calculated privacy budget $\epsilon$ (see Theorem.~\ref{thm:dg_privacy} and Algorithm~\ref{algo:DPfedpost}). 

The parameter $\delta$ is usually set to the inverse of the size of the population that the privacy mechanism is applied on, here 10 clients. However, $\delta=10^{-1}$, which represents the probability of DP being broken, is too large for any tight theoretical guarantee. Therefore, we set $\delta=10^{-3}$ for all our calculations. 

We also fix the number of communication rounds to $T_{cl}=50$ since in the case of StdFed this seems to be enough for all the models and datasets to achieve comparable performance to the centralized setting. 

When choosing the client sampling probability $C$, we face a privacy/utility trade-off: The higher the sampling probability, the better the utility of the final model and the higher the value of the privacy budget $\epsilon$. We choose $C=0.5$ as it is the lowest value that results in comparable utility between the StdFed and centralized models. 

The sensitivity $S$ is chosen during the initialization period in collaboration with all the participants. We pick the median value of the clients updates as our clipping threshold $S$.

The noise multiplier $\sigma$ has a significant impact on the performance of the models as well the calculated privacy budget $\epsilon$ and there is a trade-off between these two: the tighter the privacy guarantee, the worse the utility of the model. Hence we are looking for a reasonable operation point where there is a balance between the value of $\epsilon$ and utility. We pick two different values of $\sigma={2,3}$ which result in $\epsilon={5.4, 8.9}$, respectively~\footnote{Given a fixed value of $\delta$, $\varepsilon$ is computed numerically  as in \cite{Abadi,MironovTZ19}. (see Section~\ref{thm:dg_privacy} for more details.}.

\textbf{DPFed to DPFed-post}: When applying the post-processing step on the client-level DP (DPFed-post Algorithm~\ref{algo:DPfedpost}), the value of parameter $P$ also needs to be fixed. This parameter represents the clipping threshold for the average update vector. To find a suitable $P$ value, we examined $P=\{1S, 2S, 3S, 4S, 5S\}$ where $S$ is the parameter update clipping threshold from the previous step. A choice of $2S$ or $3S$ gave consistent improvement across all investigated models and datsets. 
\begin{table*}\footnotesize
\floatconts
  {tab:gbsg}
  {\caption{GBSG dataset}}
    {\scalebox{1.0}{\begin{tabular}{c|c|cc|cc|cc}
    \toprule
    &&\multicolumn{2}{c|}{}&\multicolumn{2}{c|}{}&\multicolumn{2}{c}{}\\
    &&\multicolumn{2}{c|}{Non-Private}&\multicolumn{2}{c|}{$(\epsilon=5.4, \delta=10^{-3})$}&\multicolumn{2}{c}{$(\epsilon=8.9, \delta=10^{-3})$} \\ 
    Metric&Model&Centralized& StdFed& DPFed & $\text{DPFed}_{\text{post}}$ &DPFed  &$\text{DPFed}_{\text{post}}$\\
    \hline
         &DeepHit& $0.66\pm0.02$& $0.67\pm0.02$& $0.47\pm0.03$&\bm{$0.56\pm0.04$}&$0.54\pm0.03$&\bm{$0.59\pm0.03$} \\
      C-index  &CoxPH&$0.66\pm0.01$& $0.67\pm0.03$& $0.45\pm0.70$&\bm{$0.62\pm0.02$}&$0.47\pm0.05$&\bm{$0.64\pm0.03$}\\
         $\uparrow$&CoxCC&$0.63\pm0.02$& $0.68\pm0.01$&$0.58\pm0.05$&\bm{$0.61\pm0.02$}&$0.62\pm0.02$&\bm{$0.64\pm0.03$}\\
         &CoxTime&$0.64\pm0.01$&$0.67\pm0.01$&$0.57\pm0.07$&\bm{$0.62\pm0.02$}&$0.61\pm0.03$&\bm{$0.63\pm0.02$}\\
         \hline
         &DeepHit&$0.18\pm0.01$&$0.21\pm0.01$&$0.29\pm0.05$&\bm{$0.20\pm0.01$}&$0.21\pm0.01$&\bm{$0.19\pm0.01$}\\
      IBS  &CoxPH&$0.18\pm0.01$&$0.18\pm0.01$&$0.36\pm0.08$&\bm{$0.20\pm0.01$}&$0.29\pm0.08$&\bm{$0.19\pm0.01$}\\
        $\downarrow$ &CoxCC&$0.19\pm0.01$&$0.18\pm0.01$&$0.30\pm0.06$&\bm{$0.19\pm0.01$}&$0.20\pm0.01$&\bm{$0.18\pm0.01$}\\
         &CoxTime&$0.18\pm0.01$&$0.18\pm0.01$&$0.26\pm0.05$&\bm{$0.19\pm0.01$}&$0.20\pm0.01$&\bm{$0.19\pm0.01$}\\
         \hline
         &DeepHit&$0.54\pm0.01$&$0.62\pm0.03$&$1.73\pm0.80$&\bm{$0.57\pm0.01$}&$0.63\pm0.06$&\bm{$0.56\pm0.01$}\\
        NIBLL &CoxPH&$0.53\pm0.01$&$0.53\pm0.01$&$1.70\pm0.90$&\bm{$0.56\pm0.02$}&$1.51\pm0.90$&\bm{$0.55\pm0.03$}\\
      $\downarrow$ &CoxCC&$0.56\pm0.02$&$0.52\pm0.01$&$1.69\pm0.90$&\bm{$0.56\pm0.01$}&$0.59\pm0.03$&\bm{$0.54\pm0.01$}\\
         &CoxTime&$0.54\pm0.01$&$0.52\pm0.01$&$0.87\pm0.22$&\bm{$0.56\pm0.02$}&$0.57\pm0.01$&\bm{$0.55\pm0.01$}\\
         \bottomrule
    \end{tabular}}}
\end{table*}

\subsection{Observations}
\label{sec:observations}
The results per dataset are shown in Tables~.\ref{tab:gbsg} (and ~.\ref{tab:metabric}, ~.\ref{tab:support} in Appendix B). The Non-Private column contains the results for centralized setting and federated training of the model using the standard FedAvg Algorithm.~\ref{algo:fed}. The two other columns each compare the results when regular differentially-private federated learning is used to when our post-processing technique is applied, for each privacy regime. The standard deviations are reported over 5 random splits of the data as explained in Section.~\ref{sec:setup}. The arrows next to metrics indicate if a lower or a higher value is desired. 

The first observation is that the average performance of standard federated training of these survival models is always at a comparable level to the centralized setting. This is an important finding, showing that with federation of small datasets, over only 50 rounds of communication, acceptable utilities can be achieved.

\textbf{Investigating the effect of post-processing technique:} to be able to assess how well our proposed technique works, for each privacy regime we compare the performance of DPFed-post with DPFed. The bold numbers in the tables indicate a better relative utility for our method. We observe that for most of the models and metrics and datasets our technique results in an improved performance. In particular for GBSG dataset, we always see an improvement. From Table.~\ref{tab:data-distro} we see that GBSG has the second most number of data points and the least number of features. So this finding might be related to the fact that it is easier for models to converge for this dataset and post-processing always helps in leading the gradient in the right direction. 


To be able to measure the change in performance quantitatively, we define 
\begin{equation}
\label{eq:delta}
\Delta_{A\rightarrow B} = \pm\frac{[\text{metric}(B)-\text{metric}(A)]}{\text{metric}(A)}
\end{equation}
which measures the improvement of strategy $B$ w.r.t strategy $A$. For IBS and NIBLL a negative value of $[\text{metric}(B)-\text{metric}(A)]$ and for C-index a positive value of this parameter shows improvement. So we also multiply by $-1$  when calculating for IBS and NIBLL.

The average values of $\Delta_{\text{DPFed}\rightarrow \text{DPFed-post}}$ for each $\epsilon$ value are shown in Table.~\ref{tab:deltas}. We calculate $\Delta$ according to Equation.~\ref{eq:delta} for each dataset and metric and model and report the average over all of them. For both values of privacy budget, we observe an average improvement of performance when the post-processing step is applied: for $\epsilon=5.4$
post-processing results in $17\%$ improvement and for $\epsilon=8.9$ the average performance is improved by $12\%$. Post-processing helps more for the case of $\epsilon=5.4$ and this is expected since at lower $\epsilon$ values more DP noise is applied to the model and the convergence of the model with no regularization of the learning rate (via post-processing) becomes difficult and the final model shows noisy behavior. 

To study the effect of post-processing on reducing the standard deviations, we calculate the average relative standard deviation rstd = (std/mean) across all datasets and models and all metrics. The change of this value from DPFed to DPFed-post is shown in the second row of Table.~\ref{tab:deltas}. For both values of $\epsilon$, post-processing helps to reduce the average relative standard deviation. We observe that $\text{rstd}_{\text{Avg}}$ for DPFed is much larger for $\epsilon=5.4$ compared to $\epsilon=8.9$. This is expected since, the noise added to the model is larger in the case of $\epsilon=5.4$ and this results in the final model showing different behaviors after each round of running the experiments. Interestingly, $\text{rstd}_{\text{Avg}}$ retains a value of $0.05$ for both $\epsilon$ regimes. This indicates that our method is successful in suppressing large deviations of the model even for higher noise values.


\begin{table*}\footnotesize
    \centering
\floatconts
  {tab:deltas}
  {\caption{Quantitative comparison of the performance of DPFed and DPFed-post}}
    {\scalebox{1.0}{\begin{tabular}{c|c|c}
    \toprule
    & \multicolumn{1}{c|}{$(\epsilon=5.4, \delta=10^{-3})$}&\multicolumn{1}{c}{$(\epsilon=8.9, \delta=10^{-3})$}\\
         & DPFed$\rightarrow$DPFed-post &DPFed $\rightarrow$ DPFed-post \\
         \hline 
         $\Delta_{\text{Avg}}$&$0.17$&$0.12$\\
         $\text{rstd}_{\text{Avg}}$&$0.19 \rightarrow 0.05$&$0.07 \rightarrow 0.05$\\
         \bottomrule
    \end{tabular}}}
\end{table*}

%% file: conclusion.tex
\section{Related Work}




Cox proportional hazards model~\cite{cox1972regression} is one of the earliest and most powerful models for survival analysis. However, it makes very restrictive assumptions on the the data, such as linear dependence of the covariates and proportionality of the hazard rate over time. There has been some suggested modifications to this formulations, such as time-dependent variables~\cite{andersen1982cox, fisher1999time} or assuming a Wiener process~\cite{lee2010proportional}.

\cite{faraggi1995neural} were the first to use simple perceptron model for Cox regression. In 2018~\cite{katzman2018deepsurv} proposed DeepSurv, a deep neural network for Cox regression, and showed that their model outperforms a simple Cox proportional hazards model. ~\cite{kvamme2019time} proposed CoxTime, a modification of DeepSurv, with no proportionality assumption on the data. In a slightly different direction,~\cite{lee2018deephit} used a deep neural network with no assumption on the data to learn the survival functions.   

Due to the high sensitivity of the medical data which are used for the survival analysis, hospitals are often reluctant to share and centralize them. Fortunately, Federated Learning allows different clients to train collaboratively a common model without sharing their data. Federated learning first proposed by Google ~\cite{mcmahan2017communication, konevcny2016federated} has recently become very popular as a method of distributed training of machine learning models. It has been used for mobile keyboard prediction~\cite{hard2018federated, lim2020federated}, financial fraud detection~\cite{suzumura2019towards, yang2019ffd} and in healthcare for, e.g. patient similarity learning~\cite{lee2018privacy} and patient representation learning~\cite{liu2019two, kim2017federated}.

Although, Federated Learning is more privacy-preserving compared to its centralized counterpart, different attacks show that the adversary can still infer sensitive information by only using the model's parameters/updates. Indeed, \cite{NasrSH19,melis2019exploiting} define a membership attack that allows to infer if a particular sample is included in the dataset of a specific client. Similarly, \cite{melis2019exploiting} define an attack which aims at inferring if a group of people with a specific property, like for example skin color or ethnicity, is included in the dataset of a particular participating entity. Finally, the strongest attack is called reconstruction attack~\cite{ZhuLH19,idlg,geiping2020inverting} presented some attacks that allow an adversary to reconstruct some training data samples of some entities. A solution to prevent these attacks and provide theoretical guarantees in to use a privacy model called Differential Privacy \citep{Dwork2014book}. 

\cite{kerkoucheprivacy} uses record-level differential privacy to protect the federated training of medical data. In~\cite{kerkoucheconstrained, kerkouchecompression} client-level differential privacy has been used to protect medical data of each hospital in a federated setting. In their work, they are concerned with a binary classifier that predicts if a patient would die. They address the problem of instability of training after adding noise by reducing the sensitivity $S$ and increasing the value-to-noise levels, either by using Compressive Sensing~\cite{kerkouchecompression} or by constraining the model~\cite{kerkoucheconstrained}. 

As far as we know, our work is the first which investigates the utility-privacy tradeoff introduced by DP under such challenging but realistic settings for the privacy analysis. Indeed, we consider a scenario with a total of only 10 hospitals and where half of them participate at each round. This means that our sampling probability C is set to 0.5 and the additive noise required by DP is also very large.  

\section{Conclusion}
In this work, we tackle the challenge of using small datasets to train a differentially-private model in a federated setting. This becomes relevant when the collection of data is time-consuming and expensive, and only a few specialized data holders are interested in training a model. We propose adding a post-processing step to the popular client-level differentially private federated learning scheme. Our results indicate that this technique - which we refer to as DPFed-post - consistently improves the performances of the models and reduces the disparity in its behavior.